\documentclass[natbib]{article}

\usepackage{mathtools}
\usepackage{graphicx}

\DeclareGraphicsExtensions{.pdf,.png,.jpg}

\usepackage{times}
\usepackage{graphicx} 
\usepackage{subfigure} 
\usepackage{amsmath}
\usepackage{mathtools}
\usepackage{natbib}
\usepackage{epstopdf}
\usepackage{algorithm}
\usepackage{algorithmic}

\usepackage[dvipdfmx]{hyperref}

\usepackage{hyperref}

\def \tabref #1{~Table \ref{#1}~}

\usepackage[accepted]{icml2014}
\usepackage{listings}

\icmltitlerunning{ MILJS : Brand New JavaScript Libraries for Matrix Calculation and Machine Learning }

\begin{document} 

\twocolumn[
\icmltitle{ MILJS : Brand New JavaScript Libraries for \\
		Matrix Calculation and Machine Learning }

\icmlauthor{Ken Miura}{miura@mi.t.u-tokyo.ac.jp}
\icmlauthor{Tetsuaki Mano}{mano@mi.t.u-tokyo.ac.jp}
\icmlauthor{Atsushi Kanehira}{kanehira@mi.t.u-tokyo.ac.jp}
\icmlauthor{Yuichiro Tsuchiya}{tsuchiya@mi.t.u-tokyo.ac.jp}
\icmlauthor{Tatsuya Harada}{harada@mi.t.u-tokyo.ac.jp}
\icmladdress{ Machine Intelligence Laboratory, Department of Mechano-Informatics, The University of Tokyo }

\icmlkeywords{javascript libraries, machine learning, matrix calculation, browsers, open source}
\vskip 0.3in
]

\begin{abstract}
  MILJS is a collection of state-of-the-art, platform-independent, scalable, fast JavaScript libraries for matrix calculation and machine learning. Our core library offering a matrix calculation is called Sushi, which exhibits far better performance than any other leading machine learning libraries written in JavaScript. Especially, our matrix multiplication is 177 times faster than the fastest JavaScript benchmark. Based on Sushi, a machine learning library called Tempura is provided, which supports various algorithms widely used in machine learning research. We also provide Soba as a visualization library. The implementations of our libraries are clearly written, properly documented and thus can are easy to get started with, as long as there is a web browser. These libraries are available from {\em http://mil-tokyo.github.io/} under the MIT license.
\end{abstract}

\section{Introduction}
  Web technology has made remarkable progress in recent years. A number of sophisticated applications such as Google Maps, Gmail, and Microsoft Office Online are now comparable to traditional native applications in terms of speed. These web-based applications have become increasingly popular due to its significant advantage of platform independence. Behind this advancement is the fact that some high-speed JavaScript execution environments have been developed; for instance, Google V8 JavaScript Engine. Scientific calculation systems, however, rarely use web technology mainly because there are crucial speed limitations. Since JavaScript is an event-driven programming language with one thread, parallelization, which is a key to perform computationally-complex problems, is basically impossible.

  In terms of applications, systems with machine learning techniques have gained a great deal of attention. The prevalence of e-commerce, smart phones, and social networks accelerated the accumulation of vast amount of information, and evoked demands for analyzing them. Some web-based machine learning libraries aiming to utilize the nature of platform independence have already been developed. ConvNetJS \cite{ConvNetJS} is one such library which implements Deep Convolutional Neural Network. Although it has a cutting-edge machine learning technique, the speed makes it impractical for research use.

  Sushi tackles the speed limitation and realizes parallelization by using WebCL. We achieve matrix multiplication that is 177 times faster than the existing fastest library \cite{Sylvester} in JavaScript. Consequently, machine learning algorithms are computed within practicable amount of time.

The contributions of this paper are threefold.
\begin{itemize}
  \setlength{\itemsep}{-0.7mm}
  \item Development of  the fastest JavaScript library for matrix calculation.
  \item Implementation of a platform-independent and scalable machine learning library.
  \item Provision of a sophisticated data visualization tool.
\end{itemize}
This paper offers both a brief introduction to MILJS and a glimpse of its superior performance.

\clearpage

\section{Libraries Overview}
  As shown in Figure 1, MILJS is composed of three libraries: Sushi, Tempura, and Soba.

\vspace{-0.5mm}
\begin{figure}[!ht]
  \label{fig:pipeline}
  \begin{center}
    \includegraphics[width=\columnwidth]{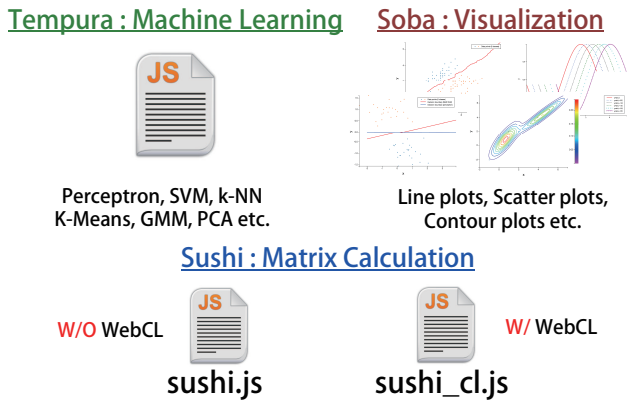}
    \caption{Overview of MILJS}
  \end{center}
\end{figure}
\vspace{-0.5mm}

\subsection{Sushi}
  Sushi is a matrix calculation library, which provides parallelized matrix calculation on multicore CPU or GPU. It uses a JavaScript binding of OpenCL called WebCL. As of January, 2015, there is no major browser which originally supports WebCL, and it is necessary to install WebCL extension on Firefox \cite{NokiaWebCL} or to use forked project from Chromium \cite{ChromiumWebCL}. Sushi thus deals with both WebCL-compliant and non-WebCL-compliant browsers. {\tt sushi.js} is implemented without using WebCL and {\tt sushi\_cl.js} is with WebCL. By loading in order, the latter overwrites some functions to use WebCL if it is available.

\subsubsection{sushi.js}
  A constructor of matrix objects and various matrix calculations using native JavaScript are implemented in {\tt sushi.js}. Elements in a Matrix are stored in the form of one-dimensional {\it Float32Array} because data on WebCL-compliant browsers are transferred between the memories of JavaScript runtime and those of OpenCL. When setting values to the array, both row-wise way and column-wise way are possible. Sushi supervises the direction by a flag, which saves the overhead of memory transportation and allows high-speed transposition.

  On the basic structure, useful functions and methods for matrix handling are implemented such as four arithmetic operations, random initialization, JSON export/import and convolution. See also Sushi API reference for details.

\subsubsection{sushi\_cl.js}
  Again, what matters in our approach is achieving the fastest matrix calculation by using WebCL. {\tt sushi\_cl.js} is a patch file which deals with that critical point and overwrites functions defined in {\tt sushi.js}. It automatically detects computing platforms and works only when WebCL is available. Below is the flow of executing codes with WebCL.

\begin{enumerate}
  \setlength{\itemsep}{-0.7mm}
  \item Allocate a WebCL object
  \item Choose platforms
  \item Choose devices
  \item Generate a context
  \item Generate a command queue
  \item Generate kernels
  \item Allocate buffers and transfer data
  \item Execute kernels
  \item Write back data
\end{enumerate}

\textbf{Step 1 :} Sushi allocates a WebCL object. WebCL is built around two types of definition either as an object or a constructor. In order to avoid inconvenience, Sushi specifies the execution environment and allocates WebCL object in both cases.

\textbf{Steps 2 and 3 :} Sushi chooses platforms and devices. The concept of platform is similar to that of device driver. It translates OpenCL codes to readable codes for each GPU or CPU. If there are several platforms available, Sushi looks up a list of priorities and picks up a high-performance platform. On common-sense grounds, GPU platforms have priority over CPU platforms to make best use of their computational ability. The selected platform sometimes provides several available devices. For example, a platform may have multiple GPUs or it may have WebCL-compliant CPU and onboard GPU. In this case, however, Sushi uses the head of these devices because of lacking sufficient information to compare their performances.

\begin{table*}[!th]
  \centering
  \caption{Overview of main algorithms included in Tempura}
  \vspace {0.3mm}
  \begin{tabular}{| c | c |}
    \hline
    Categories                    & Algorithms                            \\ \hline \hline
    Linear Models            & Linear Regression, Lasso Regression   \\
                                      & Ridge Regression, Logistic Regression \\
                                      & SGDRegressor (Perceptron, SGD-SVM)    \\ \hline
    Neighbors                & Nearest Neighbors                     \\ \hline
    Clustering               & K-means                               \\ \hline
    Mixture                 & Gaussian Mixture Model                \\ \hline
    Dimensionality Reduction & Principal Component Analysis          \\ \hline
    Cross Decomposition      & Canonical Correlation Analysis        \\
    \hline
  \end{tabular}
  \label{tab:liblist}
\end{table*}

\textbf{Steps 4, 5 and 6 :} Sushi generates a context, a command queue, and kernels. A context controls a queue and kernels to be described. Simply stated, a context is comparable to a process. After a context is created, a command queue and kernels are produced. A command queue is a queue which stores instructions; for example, execution of operation, or transfer of memory. Instructions are arbitrarily carried out in order. Sushi synchronizes the queue only when it reads and writes memory, and thus it runs another operation by native JavaScript during its free time. A kernel is a program executed in parallel as in matrix calculation. With reference to a given context, a desired number of kernels are generated. At this point, instructions are translated in accordance with specified device. Sushi handles above steps at the moment of loading libraries, and kernels are kept by a closure with respect to each method.

\textbf{Step 7, 8 and 9 :} Sushi executes kernels. Before execution, buffers for arrays of kernel arguments are allocated on the memory of GPU, and data are transferred to there. After execution, if the results are required in JavaScript code, data are written back to the memory of CPU. Note that transfers are automatically performed by Sushi. Since the cost of data transfer is a burden to fast calculation, Sushi minimizes it as much as possible. For instance, in image classification task using neural networks, only predicted class index is transferred and all other calculations are performed on GPU.

  Although we provide basic matrix calculation in {\tt sushi\_cl.js}, it is sometimes the case that developing one's own kernel solves problems more efficiently. Sushi has a convenient function for making such a kernel without being interrupted by memory management. In addition, some common kernel generators are defined. Consider the following example. When implementing an activation layer in neural networks, one needs to apply the same activation functions to all matrix elements. First, one's own kernel is created by calling {\tt mapGenerator} function as follows.
\begin{lstlisting}[basicstyle=\ttfamily\scriptsize, frame=single]
var sigmoid = Sushi.Matrix.CL.mapGenerator(
                        '1.0 / ( exp(-a[i]) + 1.0 )' );
\end{lstlisting}
The output function is able to handle Sushi.Matrix object directly.
\begin{lstlisting}[basicstyle=\ttfamily\scriptsize, frame=single]
var output = sigmoid( input );
\end{lstlisting}
With just a few lines of codes, Sushi realizes very fast matrix calculation using GPU. See also Sushi API reference for details of our WebCL-compliant funcions and methods.

\subsection{Tempura}
  Tempura is a machine learning library written in JavaScript and built on Sushi. Tempura provides efficient, simple and powerful tools for data analysis. It offers not only programming-based plain interfaces but also Soba-based sophisticated visualization, which is helpful for research use. Machine learning algorithms take one of the following types: {\tt Linear Models, Neighbors, Clustering, Mixture, Dimensionality Reduction, Cross Decomposition}. Major algorithms have already been coded, and others are now being developed. Some utility classes make it easy to load data. All algorithms are available on the API documentation of our website \footnote{http://mil-tokyo.github.io/tempura/}. An overview of main algorithms is listed in \tabref{tab:liblist}.

\subsection{Soba}
Soba is a 2D plotting library integrated with Sushi written in JavaScript. It is easy to use, works in web browsers, and coordinates with other two libraries. Moreover, Soba designs stylish figures, which offers more choices for the application of various statistical analysis methods. At this point, we give priority to the development of some frequently used visualizing tools for machine learning such as line plots, scatter plots, and contour plots.

\section{Experiments}

\begin{table*}[!th]
  \centering
  \caption{Four tasks used in the speed comparison with existing libraries}
  \begin{tabular}{| l | c |}
    \hline
    Task 1 & Addition of two 1000-by-1000 matrices \\ \hline
    Task 2 & Multiplication of a 1000-by-100 matrix and a 100-by-10 matrix \\ \hline
    Task 3 & Multiplication of a 1000-by-100 matrix and a 100-by-1000 matrix \\ \hline
    Task 4 & A Collection of operations. \\
                    & A: Multiplication of a 200-by-500 matrix and a 500-by-200 matrix \\
                    & B: Addition of the result from 4A and a 200-by-1 matrix \\　
                    & C: Multiplication of transpose of a 200-by-500 matrix and a 200-by-50 matrix \\
    \hline
  \end{tabular}
  \label{tab:tasks}
\end{table*}

\begin{table*}[!th]
  \centering
  \caption{Comparison of average runing times (mm) of four tasks.}
  \begin{tabular}{|l | l| c| c| c| c|}
    \hline
    Environment & Libraries       & Task1 & Task2 & Task3 & Task4 \\ \hline \hline
    Firefox              & Sushi with WebCL    & \textbf{10.2}  & \textbf{8.0}   & \underline{\textbf{22.0}}  & \underline{\textbf{21.2}}  \\
                         & Sushi without WebCL & \underline{2.6} & \underline{3.0}            & 228.8          & 23.8           \\
                         & Sylvester                & 46.0           & 3.6            & 623.6          & 52.2           \\
                         & math.js                  & 573.6          & 731.8          & N/A            & N/A            \\
                         & Closure Library          & 46.4           & 73.2           & N/A            & N/A            \\ \hline
    node.js              & Sushi with WebCL    & \underline{\textbf{2.0}}   & \underline{\textbf{0.6}}   & \underline{\textbf{16.4}}  & \underline{\textbf{2.4}}   \\
                         & Sushi without WebCL & 21.6           & 28.0           & 2530.0         & 194.4          \\
                         & Sylvester                & 107.2          & 17.2           & 2910.0         & 166.4          \\
                         & math.js                  & 209.2          & 130.4          & N/A            & N/A            \\
                         & Closure Library          & 148.0          & 272.2          & N/A            & N/A            \\
    \hline
  \end{tabular}
  \label{tab:comparison}
\end{table*}

  To demonstrate the efficiency of matrix calculation implemented in Sushi, we present a comparison of average running times of several tasks with existing libraries listed below.
\begin{itemize}
  \setlength{\itemsep}{-0.7mm}
  \item Sushi with WebCL
  \item Sushi without WebCL
  \item Sylvester \cite{Sylvester}
  \item math.js \cite{MathJS}
  \item Closure Library \cite{ClosureLib}
\end{itemize}
  Sylvester is the most popular JavaScript matrix calculation library since 2007, and it is the fastest among all other existing libraries. math.js is an integration solution for mathematics published recently. Closure Library is a set of JavaScript libraries developed by Google. Four tasks in \tabref{tab:tasks} are conducted in total. Task 1, 2 and 3 are simple matrix calculations, and Task 4 replicates the procedure of forward-propagation and that of back-propagation in Perceptron.

  Using MacBook Pro containing an Intel Core i5 processor clocked at 2.4 GHz and 8 GB of RAM, we conduct these tasks five times in Firefox or in node.js, which is a multi-platform runtime environment for server-side and networking JavaScript applications. The average running times are summarized in \tabref{tab:comparison}. The value of N/A indicates the computation takes too long to measure.

  Sushi completes all the tasks much faster than math.js and Closure Library regardless of the usage of WebCL. Sushi is superior to Sylvester in every task on node.js and shows better performance on Firefox except for Task 2. It is particularly worth noting that Sushi terminates Task 3 on node.js 177 times faster than Sylvester. Since the computational overhead with WebCL becomes relatively less for computationally-intensive tasks, Sushi achieves outstanding performance in Task 3 and 4.

\section{Conclusion and Future Plans}
  In this paper, we address the problem of speed limitation in JavaScript, and introduce Sushi, which achieves an exceptional performance in matrix calculation by utilizing WebCL. Seeking for more convenience, coding new functions and promoting algorithm efficiency are in progress. Based on Sushi, powerful machine learning library with well-looking visualizer is provided. Further implementations of popular techniques are in hand as well.


\bibliographystyle{icml2014}
\bibliography{paper}

\clearpage

\appendix

\def\thesection{Appendix}
\section{Demos}
With the use of Tempura and Soba, several visualization outcomes of popular machine learning techniques are described below. In these codes, {\tt \$M} equals {\tt Sushi.Matrix}, and {\tt \$S} equals {\tt Tempura.Utils.Statistics}.

\begin{lstlisting}[basicstyle=\ttfamily\scriptsize, frame=single]
// fit
var gmm = new Tempura.Mixture.GMM(2,100,0.0000001);
gmm.fit(X);

// plot
var x = $M.getCol(X,0);
var y = $M.getCol(X,1);
plt.contourDesicionFunction(
  $M.min(x)-1, $M.max(x)+1, $M.min(y)-1, $M.max(y)+1,
    function(x,y){
      var datum = $M.fromArray([[x],[y]]);
      return gmm.score(datum);
  }
);

// draw
plt.scatter(x,y);
plt.xlabel('x'); plt.ylabel('y');
plt.colorbar();
plt.show();
\end{lstlisting}

\begin{figure}[ht]
  \begin{center}
    \includegraphics[width=7.0cm]{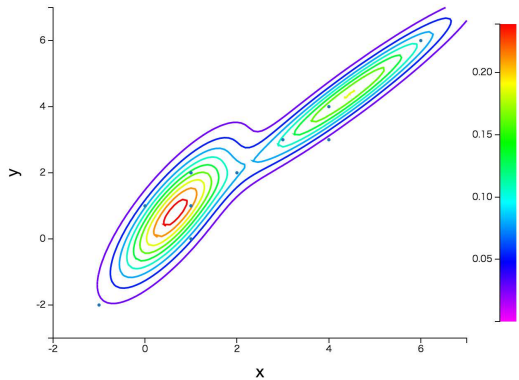}
    \caption{Demo: Gaussian Mixture Model}
  \end{center}
\end{figure}

\begin{lstlisting}[basicstyle=\ttfamily\scriptsize, frame=single]
// fit
var k=3;
var clf = new Tempura.Neighbors.KNeighborsClassifier(
                                  {n_neighbors:k});
clf.fit(samples,labels);

// plot
var x = $M.getCol(samples,0);
var y = $M.getCol(samples,1);
var color = labels.t(); plt.scatter(x,y,color);

// draw
plt.contourDesicionFunction(
  $M.min(x)-1,$M.max(x)+1,$M.min(y)-1,$M.max(y)+1,
  {levels:[1.5]}, function(x,y){
    var datum = $M.fromArray([[x,y]]));
    return clf.predict(datum.get(0,0));
  }
);
plt.xlabel('x'); plt.ylabel('y');
plt.legend(['Datapoints(2classes)']);
plt.show();
\end{lstlisting}

\begin{figure}[ht]
  \begin{center}
    \includegraphics[width=7.0cm]{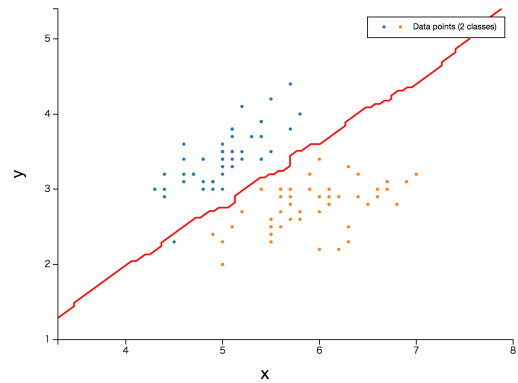}
    \caption{Demo: k-nearest-neighbors}
  \end{center}
\end{figure}

\begin{lstlisting}[basicstyle=\ttfamily\scriptsize, frame=single]
// fit
var per = new Tempura.LinearModel.SGDRegressor(
       {algorithm:'perceptron',aver:false,lambda:0.0});
var svm = new Tempura.LinearModel.SGDRegressor(
                        {algorithm:'sgdsvm'});
per.fit(samples,labels); svm.fit(samples,labels);

// plot
var x = $M.getCol(X,0); var y = $M.getCol(X,1);
var color = $M.getCol(labels,0);
plt.scatter(x,y,color);

// draw
plt.contourDesicionFunction(-2,4,-2,2,
  {levels:[0],colors:'r',linestyles:['solid']},
  function(x,y){
    var datum = $M.fromArray([[x,y]]))
    return svm.predict(datum).get(0,0);
  }
);
plt.contourDesicionFunction(-2,4,-2,2,
  {levels:[0],colors:'b',linestyles:['solid']},
  function(x,y){
    var datum = $M.fromArray([[x,y]]))
    return per.predict(datum).get(0,0);
});
plt.xlabel('x'); plt.ylabel('y');
plt.legend(['Datapoints (2 classes)',
  'Decision boundary (SGD SVM)',
  'Decision boundary (perceptron)']
);
plt.show();
\end{lstlisting}

\vspace {-4.0mm}

\begin{figure}[ht]
  \begin{center}
    \includegraphics[width=7.0cm]{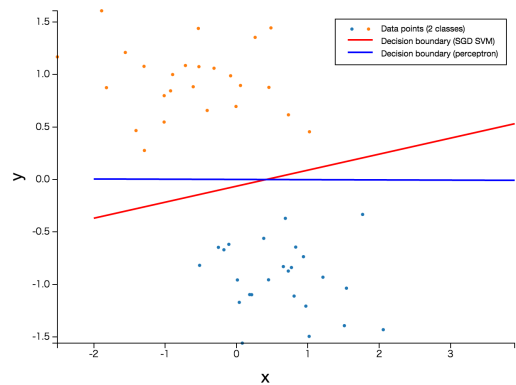}
    \caption{Demo: SGD-SVM and Perceptron}
  \end{center}
\end{figure}

\end{document}